# Experimental validation of UAV search and detection system in real wilderness environment

Stella Dumenčić[a], Luka Lanča[a], Karlo Jakac[a] and Stefan Ivić[a*]

[a]*Faculty of Engineering, University of Rijeka, Vukovarska 58, 51 000 Rijeka, Croatia*

Correspondence: stefan.ivic@uniri.hr

## Abstract

Search and rescue (SAR) missions require reliable search methods to locate survivors, especially in challenging or inaccessible environments. This is why introducing unmanned aerial vehicles (UAVs) can be of great help to enhance the efficiency of SAR missions while simultaneously increasing the safety of everyone involved in the mission. Motivated by this, we design and experiment with autonomous UAV search for humans in a Mediterranean karst environment. The UAVs are directed using Heat equation-driven area coverage (HEDAC) ergodic control method according to known probability density and detection function. The implemented sensing framework consists of a probabilistic search model, motion control system, and computer vision object detection. It enables calculation of the probability of the target being detected in the SAR mission, and this paper focuses on experimental validation of proposed probabilistic framework and UAV control. The uniform probability density to ensure the even probability of finding the targets in the desired search area is achieved by assigning suitably thought-out tasks to 78 volunteers. The detection model is based on YOLO and trained with a previously collected ortho-photo image database. The experimental search is carefully planned and conducted, while as many parameters as possible are recorded. The thorough analysis consists of the motion control system, object detection, and the search validation. The assessment of the detection and search performance provides strong indication that the designed detection model in the UAV control algorithm is aligned with real-world results.

Keywords: UAV, search and rescue, experiment, YOLO, HEDAC

# 1. Introduction

Unmanned Aerial Vehicles (UAVs) have emerged as efficient tools in Search and Rescue (SAR) missions due to their ability to rapidly access remote, often challenging or inaccessible areas enhancing the speed of locating individuals in distress, especially in situations involving natural hazards and risks. A critical aspect of this capability is person detection from aerial imagery. However, obtaining images for research purposes for this application is a challenging task since it requires access to real-world SAR scenarios, which have the focus set on the mission instead of collecting data for research purposes and often require complete anonymity. Experiments conducted in controlled and monitored conditions can therefore serve as a valuable alternative for generating datasets and advancing research of SAR missions. Additionally, existing datasets and algorithms often fall short in addressing the unique challenges posed by SAR scenarios, especially when detecting small-sized objects, such as individuals captured from a top-down perspective (Hong et al., 2021). Unlike typical object detection datasets, the visual representation of people in such images deviates from conventional forms, emphasizing the need for specialized datasets tailored to this task (Akshatha et al., 2023).

Beyond computer vision detection, effective motion control is essential for UAVs to systematically and efficiently survey target areas. Ergodic search algorithms, such as the Heat equation-driven area coverage (HEDAC) method (Ivic, 2020) used in this study, enhance the search performance by distributing the search efforts proportional to the likelihood of locating a target showing significant efficiency in SAR missions. However, implementing such strategies in SAR mission environments introduces challenges, including obstacle avoidance, real-time communication, and coordinating multiple UAVs. This is why the robustness of the motion control system is of utmost importance. The UAV search framework (Lanča et al., 2024) used in this study, utilizes the ergodic HEDAC motion control system in combination with Model Predictive Control (MPC) to efficiently search a large area while performing aerial imagery. The underlying sensing model is based on the performance of the used YOLOv8 object detection model, meaning the UAV's motion is influenced by both the target probability density function and the detection model. As UAVs adjust their flight height while searching complex terrains, the performance of YOLOv8 varies based on the flight height impacting the ground sampling distance (GSD) (Petso et al., 2021), (Qingqing et al., 2020). Since existing person detection models often do not provide performance metrics across extensive flight height and GSD ranges, the existing pretrained model was additionally trained on our initial experiment data to fill this gap.

Despite recent advancements, SAR applications still face challenges in both, computer vision object detection and UAV motion control. Detection algorithms must contend with varying environmental conditions, occlusions, and the inherently low resolution of humans in aerial imagery. Meanwhile, motion control demands adaptive strategies capable of balancing efficiency and reliability in high-stakes operations.

Motivated by these challenges, we conducted a simulated SAR scenario in controlled and monitored conditions on the Učka mountain in Croatia in 2024 resulting in the search and motion control experiment validation, as well as a dataset designed for detecting individuals in SAR scenarios. This dataset offers different image contexts, scales, and orientations reflective of real-world conditions enhancing the robustness and applicability of detection algorithms that will be valuable for future SAR research.

The paper is structured as follows: The Introduction 1 presents the problem and motivation of using object detection in combination with UAV motion control. In section 2 the related work is presented. In section 3 the motion control algorithm and the detection methodology are described. In section 4 the experiment setup is described. The results are shown in section 5. In 6 the presented methodology and its usage is discussed. The paper is concluded in section 7.

## 2. Literature overview

In the following section the utilization of UAVs in SAR missions is explored focusing on the ergodic motion control and other strategies for efficient search area coverage, as well as the usage of object detection models to help detect individuals in distress.

*2.1 UAV in SAR missions*

A detailed survey on the usage of UAVs in SAR missions is presented in (Lyu et al., 2023) giving an overview of different types of UAVs that can be used in SAR missions, as well as different operational scenarios of the UAVs in times of disasters. The advantages that UAVs offer, such as accessing inaccessible and often dangerous areas include improved safety for human resources, cost-effective operations, and faster data collection including the ability to gather high-resolution imagery or data used for research and monitoring. Equipped with advanced sensors, such as cameras including thermal cameras, multispectral cameras, and light detection and ranging (LiDAR), UAVs can be used to detect human body heat, identify structural damages, and map complex terrains. This is extremely important in situations of natural hazards and risk such as avalanches (Silvagni et al., 2017), (Bejiga et al., 2017), albrigtsen2016application} or earthquakes (Qi et al., 2016), (Calamoneri et al., 2022), (Nedjati et al., 2016). Additionally, UAVs are increasingly being integrated with communication systems and payload delivery mechanisms to expand their functional roles in SAR missions (Doherty & Rudol, 2007). This can include delivering critical supplies, such as medical kits, food, and water, to individuals in inaccessible locations. UAVs can also act as airborne relay stations as detailed in (WU et al., 2019). This method can be used to establish communication links in areas where conventional networks are disrupted, ensuring coordination among rescue teams.

The effectiveness of UAVs in SAR missions is further enhanced by advancements in motion control and object detection technologies, which play a crucial role in enabling efficient search missions while navigating complex environments and identifying targets. Motion control systems enable UAVs to maintain stability and manoeuvrability in challenging conditions, such as strong winds or obstructed terrains, ensuring reliable performance during missions, as well as effective path planning to search the target area effectively. Similarly, object detection algorithms allow UAVs to identify search targets, monitor hazards, and detect critical infrastructure, facilitating decision-making processes. This can be done either on-board the UAV or offline on a ground-based workstation. On-board processing enables real-time detection, providing immediate results but demanding significant computational resources, which reduces the battery life and limits the UAV's operational duration. On the other hand, offline processing involves analyzing captured images on a dedicated workstation with mostly better computing power, allowing for faster and more efficient processing while conserving UAV battery life, thereby extending the overall search duration.

*2.2 UAV search and ergodic motion control*

The ability to effectively control the motion of UAVs is crucial in a variety of applications, especially in situations that depend on the control efficiency such as SAR missions. In these operations, UAVs can be deployed either independently or in coordination with ground search teams to increase the search efforts as discussed in (Goodrich et al., 2008). Additionally, various search strategies have been explored to optimize UAV motion control such as the methods presented in (Lin & Goodrich, 2009) that use straight paths in combination with 90° turns to enhance the coverage in SAR scenarios.

Ergodic motion control has emerged as a promising solution due to its capability to efficiently guide the UAVs over a defined area. By leveraging the principles of ergodicity, this approach ensures that the spatial distribution of the UAV's trajectory aligns with the probability distribution of the target's presence, in particular, areas within the search domain. The benefits of using ergodic search are presented in (Miller et al., 2015) suggesting the robustness of the method in different conditions and uncertainties. This has led to multiple ergodic motion control systems being developed. The three widely recognized approaches for controlling single or multi-agent systems in ergodic exploration are HEDAC, MPC, and Spectral Multiscale Coverage (SMC).

The HEDAC method (Ivic et al., 2016) is based on the heat equation used to create a potential field enabling efficient directing of either one or multiple agents. The HEDAC method was later improved by incorporating agent sensing and detection (Ivic,

2020). In (Ivić et al., 2022), the Finite Element Method (FEM) was employed to solve the fundamental heat equation, enhancing its ability to handle irregularly shaped domains and inter-domain obstacles without increasing computational resource needs.

MPC, also known as Receding Horizon Control (RHC), is employed to generate trajectories by optimizing a specific objective within defined constraints. In (Bircher et al., 2018), (Mavrommati et al., 2017), the MPC approach was applied to path planning in a 3D search space for both known and unknown environments, demonstrating strong scalability in the experimental validation.

SMC, introduced in (Mathew & Mezić, 2011), leverages the difference between the desired and actual trajectories to create multi-agent paths. In (Hubenko et al., 2011), the Neyman-Pearson lemma was incorporated into this method for a 2D coverage task, leading to the development of Multiscale Adaptive Search (MAS), which was experimentally tested with a single UAV in (Mathew et al., 2013), .

This study presents a comprehensive experimental validation presented in (Lanča et al., 2024) using the HEDAC algorithm for coverage control and potential field generation, integrating it with MPC to enhance the motion control by optimizing flight height enabling the control strategy to improve overall system performance and flight efficiency in uneven environments.

*2.3 UAV images object detection*

Even though recent advancements in computer vision algorithms have proven highly beneficial in many fields, especially when large datasets are available for training and testing, the availability of large, annotated datasets for SAR-specific applications remains limited, hindering the development of more robust automated detection systems. Some examples of existing datasets include (Akshatha et al., 2023), (Zhu et al., 2021), (Barekatain et al., 2017). However, the person detection from aerial images has some specific challenges such as the top-down perspective of person objects resulting in different characteristics that the object detection model should recognize. The image quality can depend on the UAV velocity, especially in SAR missions where the trade-off between the mission speed and image quality needs to be considered. Additionally, the person objects in the image are already small-scaled, but the convolutional neural network (CNN) downsampling is reducing the feature representations even more resulting in a lack of context information. These challenges could be tackled by extending the existing number of publicly available UAV image datasets enabling the models to learn from more images containing even more different image contexts.

To effectively utilize these datasets, efficient computer vision techniques are required to detect and localize objects in UAV imagery despite their small size and complex backgrounds. Object detection plays a crucial role in this process, as it involves both identifying objects and determining their precise locations within an image. One of the most popular methods to solve this task is the You Only Look Once algorithm (YOLO) which is a one-stage detector dividing the image into a grid and predicting bounding boxes and their class probabilities enabling simultaneous estimation of localization and classification. The version used in this study is the YOLOv8 (Jocher et al., 2023) created based on incremental improvements of the earlier YOLO versions presented in (P. Jiang et al., 2022), (Terven et al., 2023), (Hussain, 2023).

The usage of YOLO for object detection in nature environments has shown promising results in applications such as wildlife monitoring (Gonzalez et al., 2016) and agricultural inspection (Messina & Modica, 2020). The application of YOLO person detection on thermal images has also been widely researched (Kristo et al., 2020), (C. Jiang et al., 2020), (Kannadaguli, 2020), (Levin et al., 2016), (Teutsch et al., 2014), (Giitsidis et al., 2015), (Yeom, 2021). Additionally, in (Yeom, 2024) the tracking of people using thermal images in simulated SAR situations is shown using YOLOv5. However, detecting small objects of interest, such as people in large-scale images, is challenging due to their small scale (Hong et al., 2021).

## 3. UAV Motion control and machine vision detection

The successful usage of autonomous UAVs in SAR missions depends on several factors such as the implemented motion control and detection. In this section, the used methodology in terms of the probabilistic model of the search, the UAV motion control using HEDAC and MPC, and the YOLO object detection model are described.

### 3.1 Probabilistic model of the search

The main objective of the conducted search is to validate the search success. To achieve this, the first step is to define the UAV's field of view (FOV) as well as the terrain model needed to determine the UAV's sensing. Since the experiment search is conducted in the mountain area, the terrain is uneven, hence the sensing may not capture the whole FOV that would be visible on even terrain. This is why the terrain data as part of the geographic information system (GIS) needs to be introduced. The terrain data was obtained by digital elevation model (DEM) files from the Copernicus database (European Union space programme, 2024). The DEM data was integrated to provide the information needed for calculating relative heights in the motion control system, as well as the possible obstacles impacting the sensing. The relative flight height, defined as the height of the UAV above the ground, is calculated using the starting point of all flights, namely 45.2368° latitude and 14.2031° longitude, and the DEM data. This calculated height is used to enable the flight height optimization and defining the no-fly safety zone. In Figure 1 it can be seen how the terrain can impact the UAV's FOV.

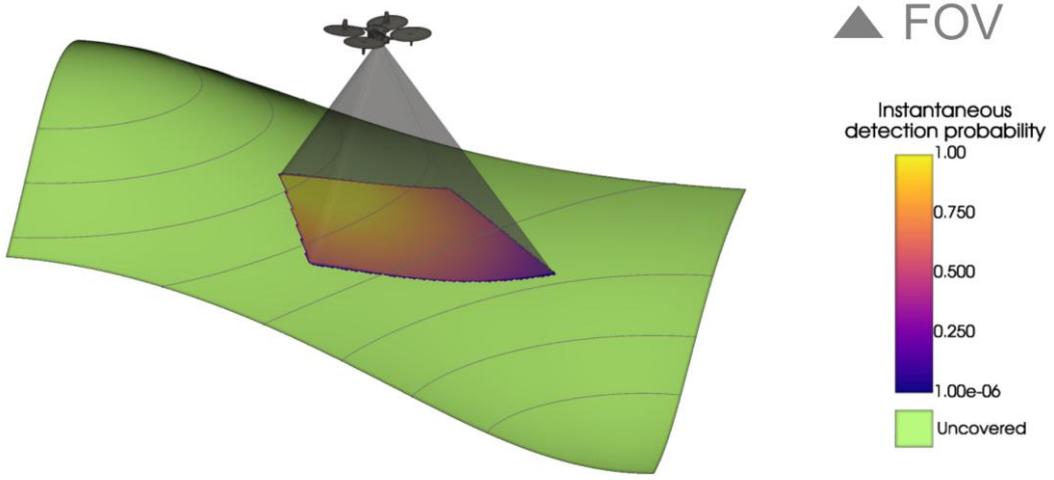

Figure 1. The FOV of a single UAV is represented by a semi-transparent pyramid. The detection probability in the UAV's FOV is shown using a gradient showing higher detection probability for points closer to the UAV's orthogonal view according to the sensing model, while the undetectable points are shown in green color.

To check if the defined point $p$ can be sensed by the camera, the point coordinates need to be transformed to local coordinates in relation to the UAV's coordinate system. By transforming the coordinates, the original x and y coordinates are used, while the z coordinate first needs to be calculated based on the UAV flight height and the terrain at the point $Z_T(x, y)$. Based on if the point is in the FOV, the detection probability $\psi$ is defined as:

$$\psi(R) = \begin{cases} \Gamma(\lor R \lor), \land\ if R \in \Omega_{FOV} \\ 0, otherwise \end{cases}, \qquad (1)$$

where $R$ is a 3D defined point relative to the UAV camera, $\Gamma$ is used to define the detection probability. For each point that is in the FOV, the detection probability is calculated by $\Gamma$, while points outside of the FOV have a 0 probability of detecting the targets.

During the whole duration of the flight, the coverage $c$ is calculated as the accumulated detection probability for all points in the domain visible from the camera's position $X$. This is accumulated to calculate the search coverage in space and time:

$$c(p, t) = \int_0^t \psi\big(R(X(t), p)\big) dt, \qquad (2)$$

The probability of undetected target presence $m$ is initially described by the probability distribution $m_0$ at $t = 0$. Over time, $m$ decreases as the agents apply their sensing effects, which are characterized by the coverage $c$. It is calculated as follows:

$$m(p,t) = m_0(p) \cdot e^{-c(p,t)} . \tag{3}$$

To calculate the overall detection probability $\eta$, the undetected targerget is integrated over the domain:

$$\eta(t) = 1 - \int_{\Omega_{2D}} m(p,t) dp . \tag{4}$$

The detection probability is the key factor analyzed in this study since it is a measure of the search effectiveness.

*3.2 UAV motion control*

The motion control system implementation used in the main experiment, was taken from (Lanča et al., 2024) and consists of the HEDAC algorithm for defining the motion control in 2D space and MPC for optimizing the flight regime in 3D space adding the height as an additional control variable, as well as the UAV velocity. Although the proposed motion control framework is designed to handle multiple UAVs, all search missions were conducted as single-agent searches.

The motion control consists of three control variables set by the motion control algorithm, namely the velocity intensity $\rho(t)$, the incline angle $\varphi(t)$, and the yaw angular velocity $\omega(t)$. Using the velocity intensity and the incline angle, the horizontal and vertical velocities are calculated. In addition, $\omega$ regulates the UAV direction in which the horizontal velocity acts. By this, the UAV state is defined using three coordinates, namely the $x$, $y$, and $z$ coordinates, as well as one orientation state.

The horizontal search control is defined by the potential field $u(p,t)$ of the search area. The potential field is guiding the UAV towards the areas that have the highest probability of containing undetected targets. It is calculated by solving the differential equation as follows:

$$\alpha \cdot \Delta u(p,t) = \beta \cdot u(p,t) - m(p,t) , \tag{5}$$

where $\alpha$ and $\beta$ are HEDAC parameters used to modify the search behavior by adjusting the smoothness and stability, and $\Delta$ is the laplace operator. Additionally, the following condition has to be met:

$$\frac{\partial u}{\partial n} = 0 \ . \qquad (6)$$

where $n$ represents the normal outward to the search domain boundary defined by $\partial \Omega_{2D}$. Based on the gradient of the potential field, the direction of the UAV needs to be adjusted. This is calculated for each control step steering on the current direction towards the wanted direction defined by the gradient. Additionally, the UAV 's maximum angular velocity is defined by the maximal turning velocity or equivalently the minimum turning radius.

To control the UAV's flight height and velocity, MPC is introduced to optimize two objectives, namely maximizing the UAV velocity, while keeping the flight height as close to the height goal defined for each flight. The first constraint that needs to be satisfied for the optimization to be feasible, is the need to fly above the minimum height representing the no-fly zone set at 35 meters above the terrain obtained by the terrain model. The no-fly zone height takes into account the tree height, possible uncertainties contained in the DEM data, as well as an additional safety factor to minimize the risk of any collision with possible obstacles. Additional constraints that need to be met are the minimum and maximum velocities defined by the UAV specifications, as well as minimum and maximum accelerations.

### *3.3 Computer vision system for human detection*

To collect all the necessary data and to test out the motion control and vision detection systems needed for the success of the main experiment, an initial experiment with 28 participants was conducted on the mountain Učka on 07.07.2024.

The initial experiment dataset was obtained by manually operated flights using DJI Matrice 210 and DJI M30T UAVs, where the obtained images have a resolution of $2970 \times 5280$ pixels for the DJI Matrice 210 and a resolution of $3000 \times 4000$ pixels for the DJI M30T. This is the initial dataset used to train the YOLO object detection model used later in the main experiment. All captured persons are manually labeled in all images in the initial dataset.

The initial dataset consists of images in combination with the corresponding labels in the YOLOv8 format representing detected individuals. The images are stored in JPG format and include metadata such as Global Positioning System (GPS) coordinates, providing valuable context for analysis. The image preprocessing includes tiling the original image into smaller parts to ensure easier YOLO training and modifying the existing labels to fit the new small-sized images.

*Image labelling*

The Computer Vision Annotation Tool (CVAT) in a local environment was used for labeling. Three independent annotators manually labeled the original-sized images identifying individuals. After the initial labeling, two independent reviewers, who were not involved in the labeling process, reviewed the annotations for accuracy and consistency. The images were labeled in an iterative process where labels were corrected to increase the accuracy. The used label format is YOLO, specifically, it was downloaded as the YOLOv8 Detection label format available in CVAT.

*Image tiling*

The process of dividing the original UAV images into smaller tiles was performed using a custom Python script. Each high-resolution image was split into 512 × 512 pixel sections, ensuring an overlap between the tiles to maintain comprehensive coverage and provide additional context for better analysis. The minimal overlap is experimentally defined as 100 px. The tiling method is shown in Figure 2.

This method enables the use of smaller square image segments preferred by the YOLO algorithm training. The newly created file names were generated based on the tile's position within the original image. This naming convention helps to ensure that each tile can be easily traced back to its location within the larger image, providing a structured approach to organizing the dataset for further processing and analysis. Lastly, the existing labels needed to be modified to fit the newly created images.

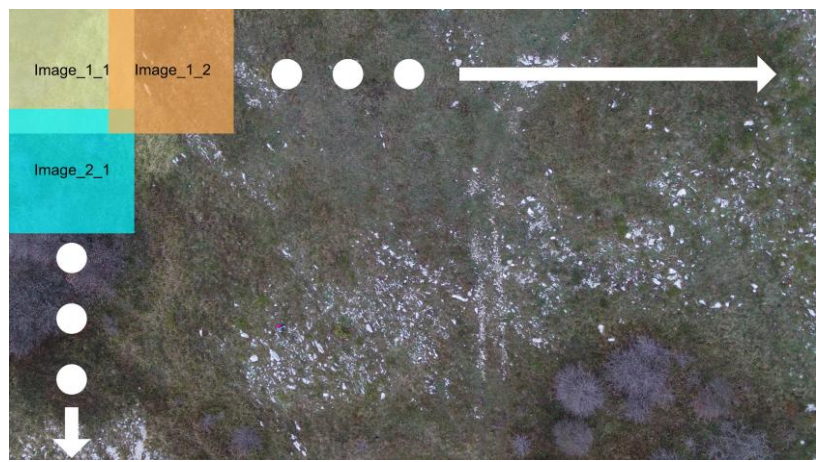

Figure 2 The tiling method used to divide original sized images into images of 512 × 512 pixels. The overlap ensures more context being available in the newly created dataset.

*Ground sampling distance*

Since different cameras were used, the images were divided into GSD groups to enable the comparison between different flight height conditions. Essentially, GSD is the actual distance in the UAV image represented by 1 px defining how much detail is captured in the image. A lower GSD means that each pixel in the UAV image is representing a smaller ground area enabling the image to show more detail, while in contrast, a higher GSD is representing a bigger area, thus having less details. Using the horizontal $h_{GSD}$ and vertical $v_{GSD}$ distance of the UAV's camera FOV, the GSD in the horizontal and vertical directions can be calculated using the relative UAV height $h$ of each image as follows:

$$h_{GSD} = 100 \cdot \frac{2 \cdot h \cdot \tan\frac{h_{FOV}}{2}}{x_{image}}, \qquad (7)$$

$$v_{GSD} = 100 \cdot \frac{2 \cdot h \cdot \tan\frac{v_{FOV}}{2}}{y_{image}}. \qquad (8)$$

Because the horizontal and vertical distance of the FOV are calculated from the aspect ratio and diagonal FOV, the vertical and horizontal GSD are the same. The GSD image groups enabled us to compare the model at different GSD intervals. The recall metric of the initial model for each GSD group is used for the motion control system. The distribution of images in height and GSD groups is shown in Figure 3.

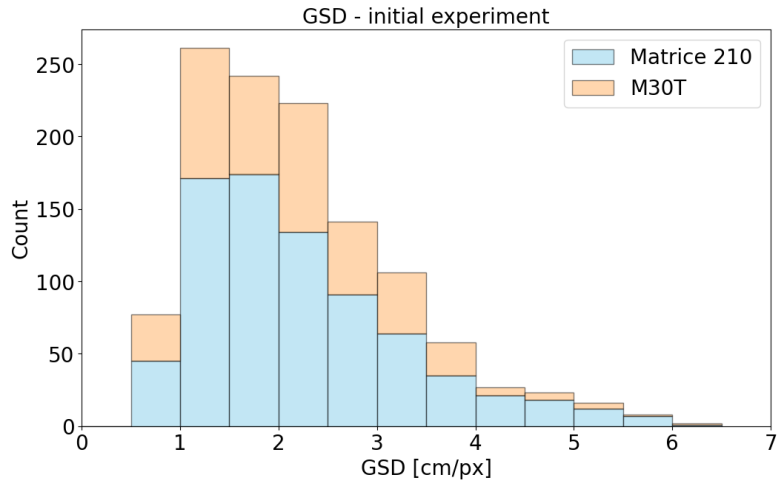

Figure 3. Image distribution based on the height and GSD. The UAV camera height is directly impacting the GSD value.

*Object detection*

The used model is YOLOv8 released in 2023 by Ultralytics and is the result of incremental improvements implemented on previous versions (YOLOv5, YOLOv6, YOLOv7, ...). The name YOLO comes from the simultaneous estimation of localization and classification that is done in one look of the images. The simplified scheme of the YOLOv8 architecture is shown in Figure 4 and consists of four main blocks: the input data, the backbone, the neck, and the head.

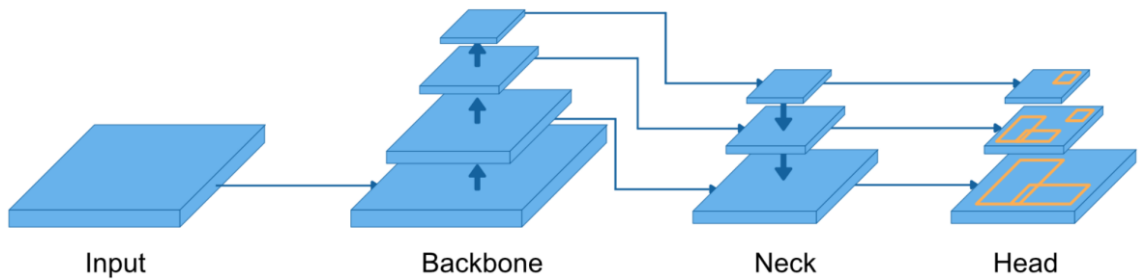

Figure 4. Simplified YOLO architecture. The input data is sent to the backbone which extracts image features by propagating it through multiple layers. The neck is enhancing the created feature maps by using different scales. The head is predicting bounding boxes.

The input data is the data provided to the model for training. The used images have a resolution of $512 \times 512$ pixels. Different augmentation methods were used to introduce new context improving the generalizability of the model. The augmentation methods included in the training process were horizontal flip, vertical flip, rotation, hue, hsv, translate, scale, mosaic, erasing, and crop fraction. Most of these methods are set as default augmentation methods. The used model is the pre-trained YOLOv8 trained on the COCO dataset.

The backbone network of the neural network is used to extract the features from the images. Based on the extracted features, object classification and localization are performed. The feature extraction is done in several layers. In our research, the originally proposed custom CSPDarknet53 is used.

In the neck block, the extracted features are aggregated to form new features from different layers of the backbone network. To aggregate features, the original PANet was used.

The head is used for suggesting anchors bounding boxes of the detected objects, in our case persons. Additionally, the head is used to estimate the percentage of certainty for detected objects. The used model head is the one presented in the original model, namely the YOLOv8 head.

*Initial detection model performance*

The sensing function is based on the YOLOv8 metrics, specifically the recall metric representing the percentage of correctly identified objects in relation to the total number of actual objects in the dataset reflecting the model's effectiveness in detecting all instances of a specific class. The recall used in the main experiment was obtained from the initial experiment.

Since the initial experiment flights were operated manually, there was no optimization of the flight height resulting in more GSD groups than in the optimized autonomous flight regime in the main experiment. The recall metric obtained by the validation on the initial dataset resulted in the recall metrics for each GSD shown in Table 1. It can be seen that the recall is generally getting lower for higher GSD intervals.

Table 1. Recall for each GSD group in the initial experiment.

| GSD | Recall |
| --- | --- |
| 0.5 - 1.0 | 0.95 |
| 1.0 - 1.5 | 0.977 |
| 1.5 - 2.0 | 0.956 |
| 2.0 - 2.5 | 0.953 |
| 2.5 - 3.0 | 0.897 |
| 3.0 - 3.5 | 0.881 |
| 3.5 - 4.0 | 0.781 |
| 4.0 - 4.5 | 0.796 |
| 4.5 - 5.0 | 0.719 |
| 5.0 - 5.5 | 0.699 |
| 5.5 - 6.0 | 0.621 |
| 6.0 - 6.5 | 0.142 |

# 4. Experiment setup

The motivation of the experiment can be divided into two main goals, namely (1) additional experimental validation of the autonomous motion control system using HEDAC and MPC and (2) creating a dataset containing people in different natural environments used in future SAR research. The experiment was conducted on the Učka mountain, Croatia on 27.10.2024. with 84 volunteers including the organizers and consists of a treasure hunt enabling the wanted motion behavior of the participants.

*4.1 Location, environment and equipment*

Učka mountain Nature Park in Croatia presents a complex and challenging environment well-suited for the evaluation of simulated UAV-based SAR operations. The area is characterized by uneven terrain, different low vegetation, and elevation variations, making it an ideal setting for assessing the capabilities of autonomous UAV systems in locating missing persons in real-world conditions.

In this experiment, two UAVs were used: DJI Matrice 210 v2 and DJI Mavic 2 Enterprise Dual. The UAVs characteristics are shown in Table 2. The specifications show the $\varphi$ parameter of optimization used for defining the UAV movement in relation to the horizontal plane, minimum and maximum horizontal and vertical velocity, minimum and maximum horizontal acceleration, maximum angular velocity, and the set MPC time steps. In Table 3 the camera specifications of three used cameras are shown.

Table 2. UAV specifications.

| UAV | Full name | Unit | Matrice 210 v2 | Mavic 2 Enterprise Dual |
|---|---|---|---|---|
| $\varphi_{min}$ | Minimum incline angle | ° | -90 | -90 |
| $\varphi_{max}$ | Maximum incline angle | ° | 90 | 90 |
| $v_{h,min}$ | Minimum horizontal velocity | m/s | 0 | 0 |
| $v_{h,max}$ | Maximum horizontal velocity | m/s | 10 | 8 |
| $v_{v,min}$ | Minimum vertical velocity | m/s | -3 | -2 |
| $v_{v,max}$ | Maximum vertical velocity | m/s | 5 | 3 |
| $a_{h,min}$ | Minimum horizontal acceleration | m/s² | -3.6 | -3.6 |
| $a_{h,max}$ | Maximum horizontal acceleration | m/s² | 2 | 2 |
| $a_{v,min}$ | Minimum vertical acceleration | m/s² | -2 | -2 |
| $a_{v,max}$ | Maximal vertical acceleration | m/s² | 2.8 | 2.8 |
| $\omega_{max}$ | Maximal angular velocity | °/s | 120 | 30 |
| N (T) | MPC horizon timesteps (duration) | - (s) | 5 (15) | 5 (15) |

Table 3. Camera specifications.

| Camera | FOV c1 [°] | FOV c2 [°] | Resolution [px] |
|---|---|---|---|
| DJI Zenmuse X5S | 39.2 | 64.7 | 5280 × 2970 |
| DJI Zenmuse Z30 | 33.9 | 56.9 | 1920 × 1080 |
| Mavic 2 Enterprise Dual built-in camera | 57.58 | 72.5 | 4056 × 3040 |

The final experiment setup is shown in Table 4. Flights 1, 2, and 3 are connected and represent a single search mission, while flights 4 and 5 each represent their own search mission resulting in a total of three search missions. All five flights were operated autonomously.

Table 4. Experiment setup settings

| Flight | 1 | 2 | 3 | 4 | 5 |
|---|---|---|---|---|---|
| Search mission | Mission 1 | | | Mission 2 | Mission 3 |
| UAV | M210 | M210 | M210 | Mavic | M210 |
| Camera | X5S | X5S | Z30 | Mavic built-in camera | X5S |
| Min/goal altitude [m] | 35/55 | 55/75 | 35/75 | 35/55 | 35/55 |
| Zone | A, B, C | A, B, C | A, B, C | B, C | A |
| Start time | 11:15 | 11:44 | 12:13 | 12:40 | 13:02 |
| End time | 11:38 | 12:07 | 12:35 | 12:55 | 13:27 |

*4.2 Design and preparation of the experiment*

Ensuring an even probability of human targets within the defined search zone involved the strategic placement of markers that needed to be found. The experiment consisted of 150 markers where 50 of them were placed in each of the three zones as shown in Figure 5 to ensure the dispersion of individuals in the search area, thereby simulating a uniform distribution of targets in the search domain. The specification of each zone is shown in Table 5. The search domain is defined using the zones and consists of either one or more zones with an offset between 50 and 100 m allowing UAVs to avoid touching the boundary of the domain. The target probability distribution function was uniform for each zone. Additionally, the sum of undetected target probability throughout the entire search domain is 1. The uniform probability inside each zone is determined by the number of people searching in that zone divided by the area of that zone.

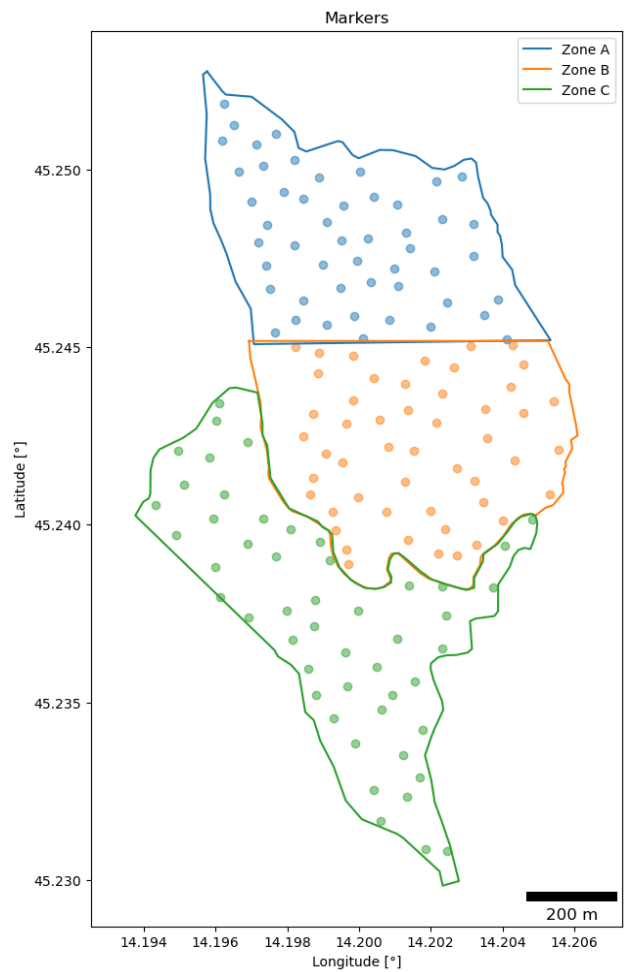

Figure 5. This figure shows three zones and their corresponding markers. Each zone contains 50 markers for the treasure hunt with unique names to identify them.

Table 5. Specification of each zone.

| Zone | Area [$m^2$] | Num. markers | Num. people |
|------|--------------|--------------|-------------|
| A    | 432 734      | 50           | 25          |
| B    | 470 233      | 50           | 27          |
| C    | 613 709      | 50           | 26          |

All participants were informed about the experiment setup and motivation. Each individual involved in the experiment provided signed consent to be photographed, with all data de-identified to protect personal information such as names. Additionally, each participant got an information flier containing all necessary information: a map with the path to the starting point, a map with defined zones, additional information, QR codes that led to the starting point, as well as QR codes that showed the position of the participant inside the zone. The available GPS data allowed the participants to track their location at all times during the experiment to ensure they remain in the assigned zone. Each participant had to fill in the name and surname, jacket color or multiple colors if they have taken off the jacket, the time at the starting point, start and end of the search inside the zone, and if they found markers, each marker should have been noted using the marker number as well as the time when the marker was found. The English version of the flier is shown in Figure 6.

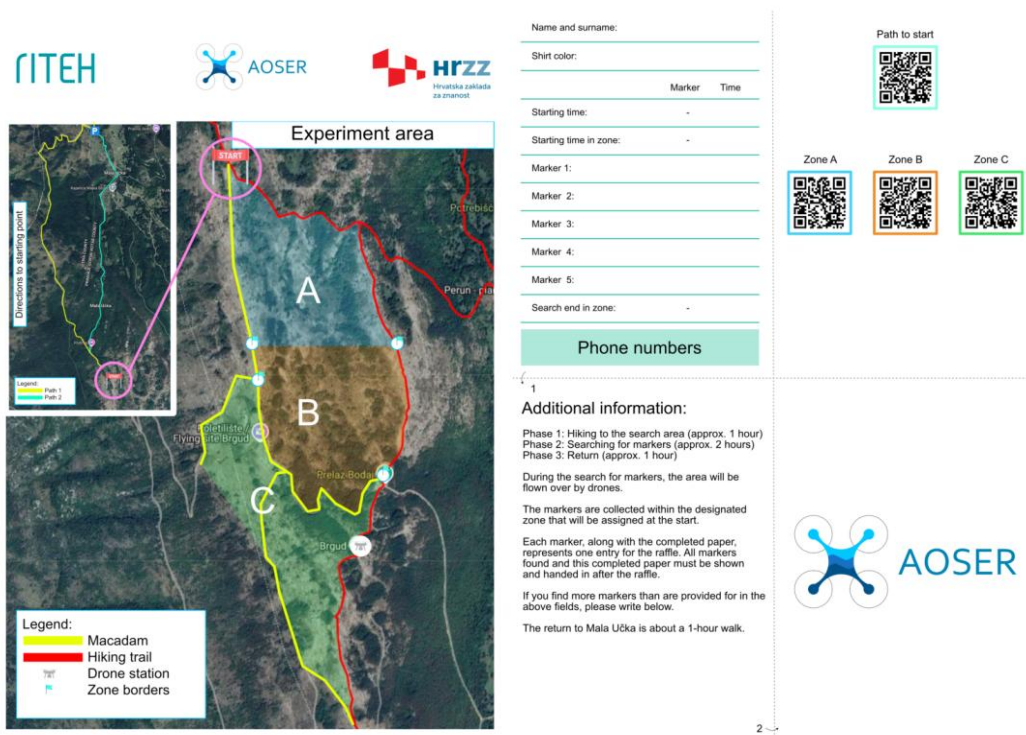

Figure 6. Information flier containing all important information for the participants as well as the log section. The original version of the flier is in Croatian, but for the purpose of showing this flier, it has been translated into English.

## 4.3 Conducting the experiment

Conducting field experiments requires detailed planning, coordination, and adaptability especially since real-world experiments introduce numerous challenges, including logistical constraints, regulatory requirements, and unpredictable environmental factors.

One of these challenges is the unpredictable weather making long-term planning for UAV-based SAR experiments challenging. While weather forecasts are monitored, sudden changes such as fog, wind, or rain can still occur, affecting UAV stability and visibility. Despite this uncertainty, extensive logistical work must be completed in advance, including inviting participants, coordinating with the nature park, obtaining flight and imaging permissions, and securing signed consent from all participants involved in the experiment. These preparations ensure regulatory compliance, operational feasibility, and safety. Even though the weather forecast seemed promising, the weather on the experiment day was cloudy and foggy.

Managing a relatively large number of participants also resulted in challenges. Some participants forgot to enter the required log data. Additionally, depending on the location inside the zones, internet connectivity issues prevented some of the participants from verifying their locations based on the GPS coordinates and maps provided in the flier, disrupting real-time decision-making resulting in some individuals straying outside of their assigned zones.

Some of the images taken on 27.10.2024. during the experiment are shown in Figure 7. Subfigures (a) and (b) show the used drones, namely Matrice 210 v2 and Mavic 2 Enterprise Dual. In (c) the home point of all flights is shown. Subfigure (d) shows the participants while giving them the introduction and explaining the experiment. In Subfigures (e) and (f) an example of a UAV image from the first mission is shown, as well as a tile of the image containing a person.

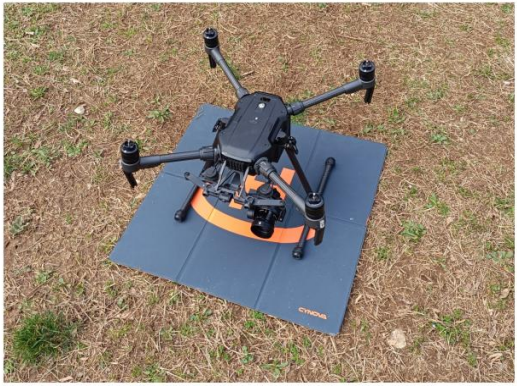
(a) Matrice 210 v2

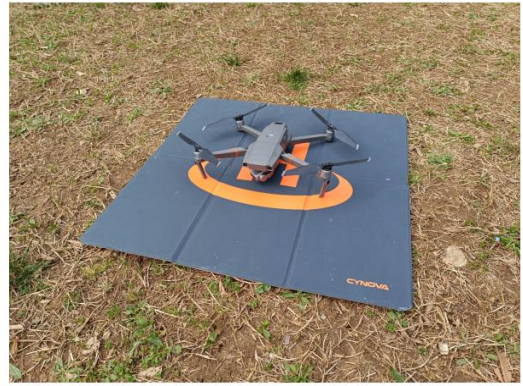
(b) Mavic 2 Enterprise Dual

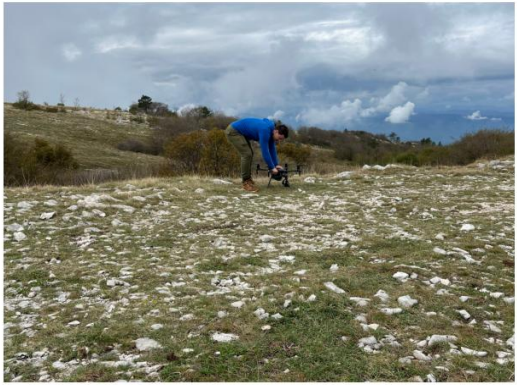
(c) Home point

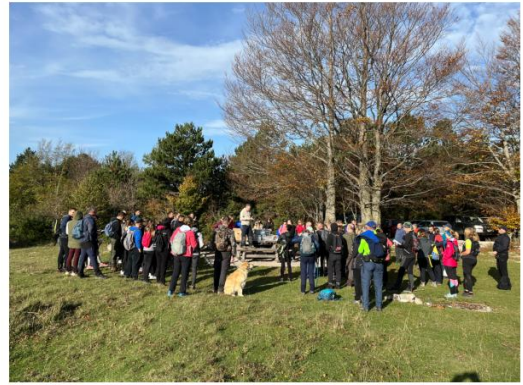
(d) Participants

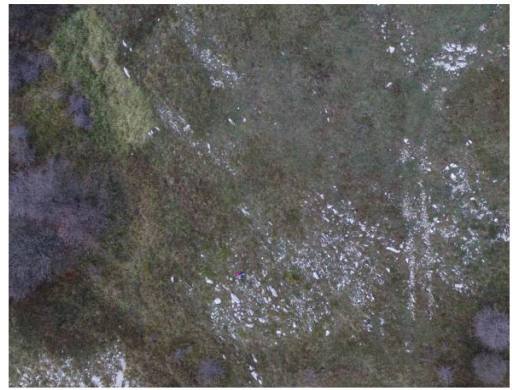
(e) UAV image

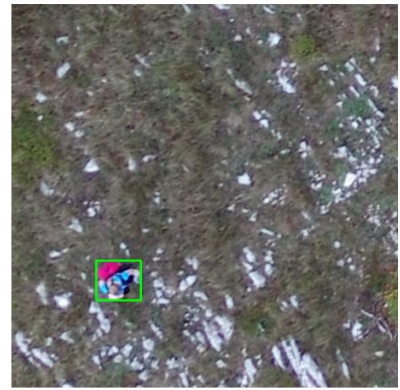
(f) Image tile

Figure 7. Scenes from conducting the experiment. Subfigures (a) and (b) show the used UAVs. Subfigure (c) shows the preparation to begin the mission at the home point of each flight. In (d) the participants are shown while the introduction speech has been given. Subfigures (e) and (f) show examples of a UAV image taken in the Mission 1 with one tile of the image.

# 5. Results

The following section presents the results of the conducted experiment consisting of the analysis of the UAV motion control, the performance of the computer vision-based human detection, and the validation of the search and detection process.

*5.1 Analysis of UAV motion control*

As mentioned in the experiment setup section, the area containing markers where people were expected to stay was divided into three zones. However, to capture the whole zone, the UAV flight zone was larger than the defined search zones. The flight trajectories of all flights during the three search missions are shown in Figure 8.

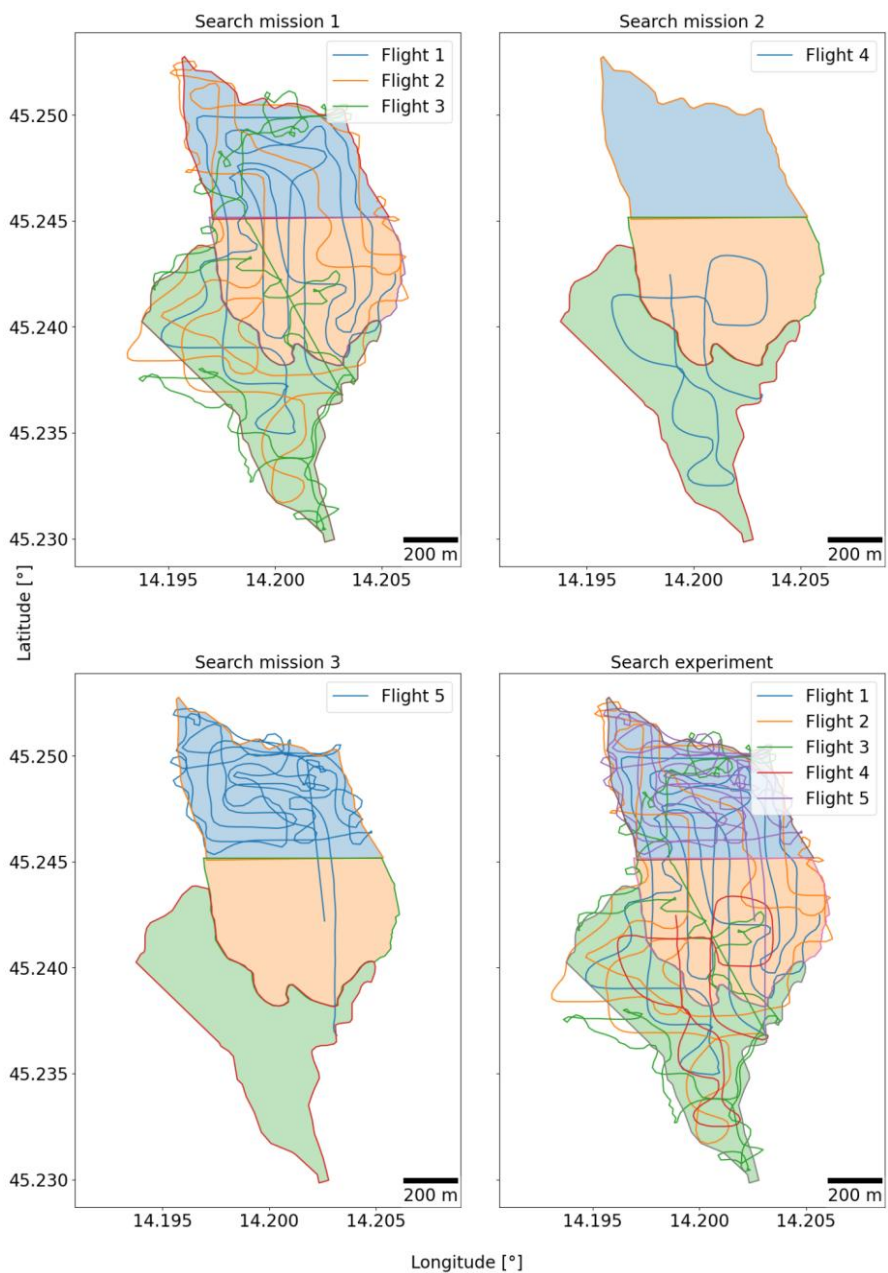

Figure 8. All flight trajectories of all search missions. (a) is showing Search mission 1 consisting of flights 1, 2, and 3. (b) is showing Search mission 2. (c) is showing Search mission 3. (d) is showing all search missions.

The resulting flight velocity, acceleration and height of the first flight in Search mission 1 is shown in Figure 9. It can be seen that the UAV's velocity and acceleration are inside the constraints defined in the MPC optimization suggesting a stable flight. The flight height is following the goal height set to 55 meters in a smoothed line allowing the UAV to optimize the velocity.

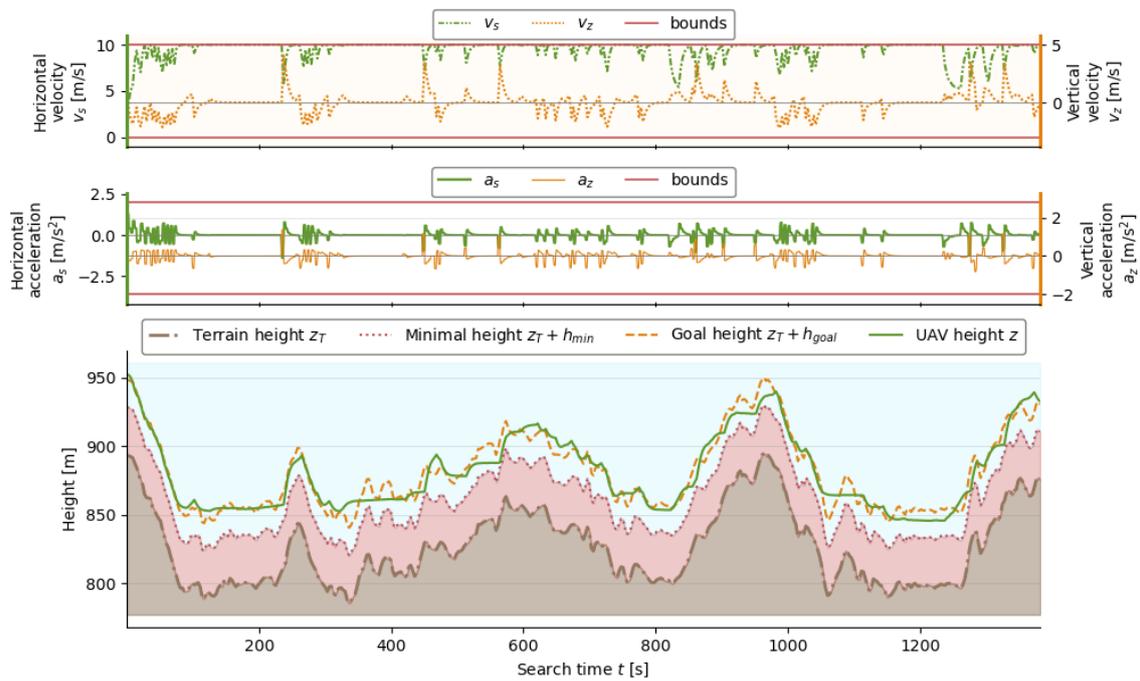

Figure 9. The first flight of Search mission 1 with a MPC horizon length of 15 seconds during a 1400 seconds flight. The UAV's velocity and acceleration is inside the set constraints. The goal flight height is set to 55 meters with the UAV maximizing the flight velocity, while minimizing the flight height resulting in a smoother line.

*5.2 Computer vision human detection*

The images resulted in most of them having no people. The number of images and number of labels in each flight is shown in Figure 10. The flight 3 which has been used to search all three zones has resulted in having the most images containing people meaning it has also the highest number of labels, averaging on two persons per image containing people. Following flight 3, flight 1 has the most images containing people, but flight 2 has more detected people.

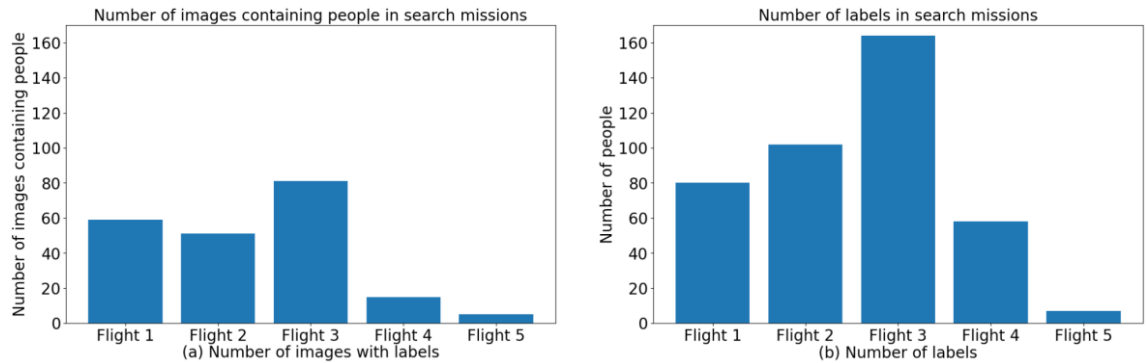

Figure 10. Number of images with labels and number of labels in all UAV flights. It can be seen that the Flight 3 contains most images as well as labels. Flight 1 contains more images with labels than Flight 2, but less labels. Even though Flight 5 consisted of only one zone, it has the lowest number of images containing people and labels.

The images with people were taken from the locations shown in Figure 11. Since the UAV flight zone is larger than the defined zones containing markers, the UAVs captured images of people that mistakenly went outside of the zone. It can be seen that in zones A and B if people went out of their zones, it is still near the zone border, while in zone C people went further outside. Most grouped images are taken at the UAV flight station as expected, since there were two flight operators at all times that are in the starting and ending images of each flight and many volunteers decided to visit the operators at the highest point of the area.

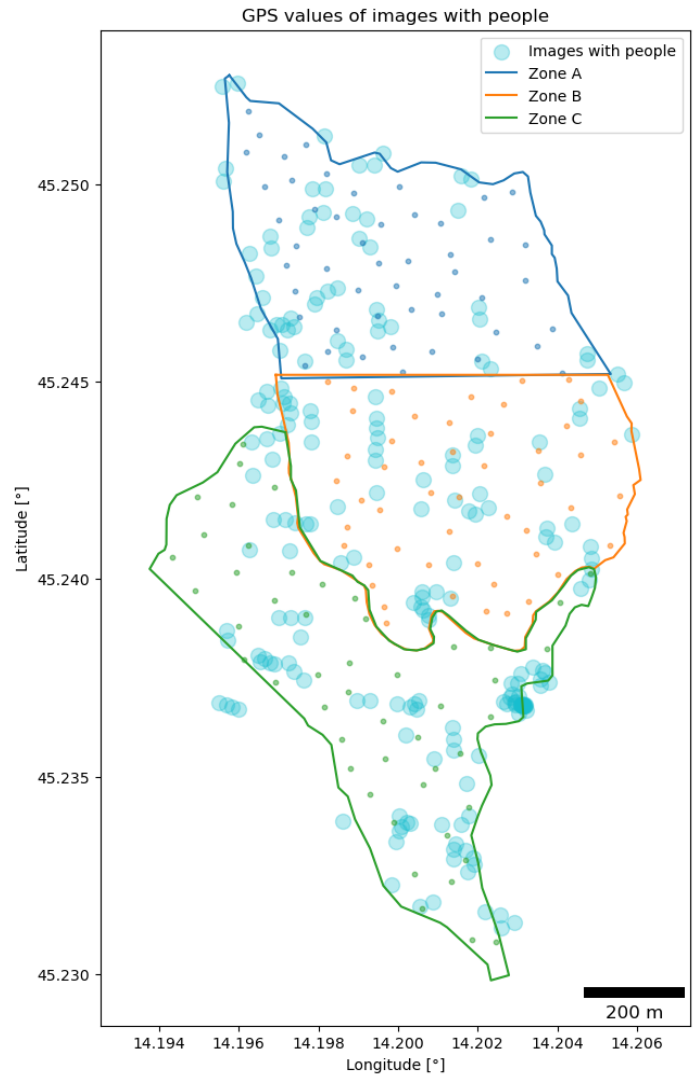

Figure 11. All images contain people and their locations. The UAV flight zone is larger than the defined zones containing markers. Most people were detected in the location of the starting point. This is expected since two UAV flight operators were in this location at all times and took images at the UAV flight start and end of each flight.

*Detection model validation*

The number of images for each GSD interval in each Search mission is shown in Figure 12. It is important to note that this Figure shows the GSD of all taken images. However, for validation, only original sized images containing people were tiled into subparts of 512 × 512 pixels since most tiles do not contain people and would still be enough to represent images with no people.

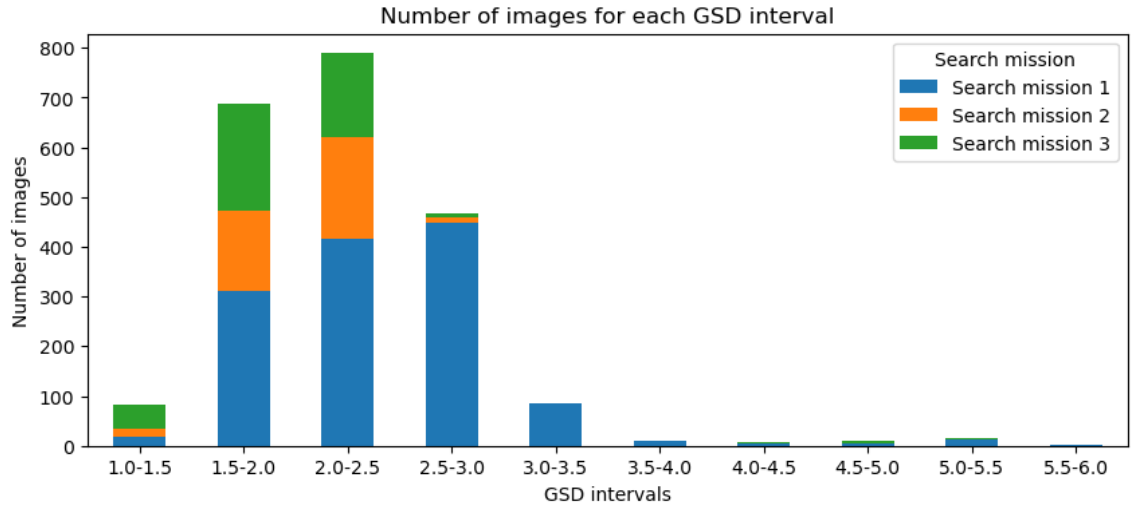

Figure 12. Number of images in each GSD interval for each Mission. Mission 1 was the longest one and consisted of three flights resulting in the largest amount of images for most GSD intervals. It is important to note that this represents all images, meaning that not only images with labels are considered, but also images without them.

The newly obtained dataset used an optimized flight regime implementation, hence having less GSD values than in the initial dataset. However, since the data in the initial dataset was taken using only one of UAVs used in this experiment, contained less people, namely about 28, and mostly having them grouped on the same walking path, therefore containing less different image contexts than in this presented experiment dataset, the recall values are lower than in the initial dataset. The resulting recall metrics for each GSD is shown in Table 6.

Table 6. Recall for each GSD group in all search missions.

| GSD | Search mission 1 | Search mission 2 | Search mission 3 |
| --- | --- | --- | --- |
| 1.0 - 1.5 | 1 | - | 1 |
| 1.5 - 2.0 | 0.71 | 0.51 | 0.67 |
| 2.0 - 2.5 | 0.68 | 0.37 | 0.38 |
| 2.5 - 3.0 | 0.8 | - | - |
| 3.0 - 3.5 | 0.37 | - | - |

The recall in each GSD interval in comparison to the initial experiment is shown in Figure 13. As mentioned earlier, since the initial experiment flights were operated manually, the flight height has not been optimized but purposely designed to gather images from a broader altitude range, resulting in more GSD ranges. On the other hand, in the main experiment, the flights were operated autonomously resulting in a stable flight height. Additionally, in the initial experiment flights, two cameras were used of which only one is used in this experiment causing new context in the ML model validation. However, it can be seen that the recall generally decreases with increasing the GSD following the trend of the initial experiment.

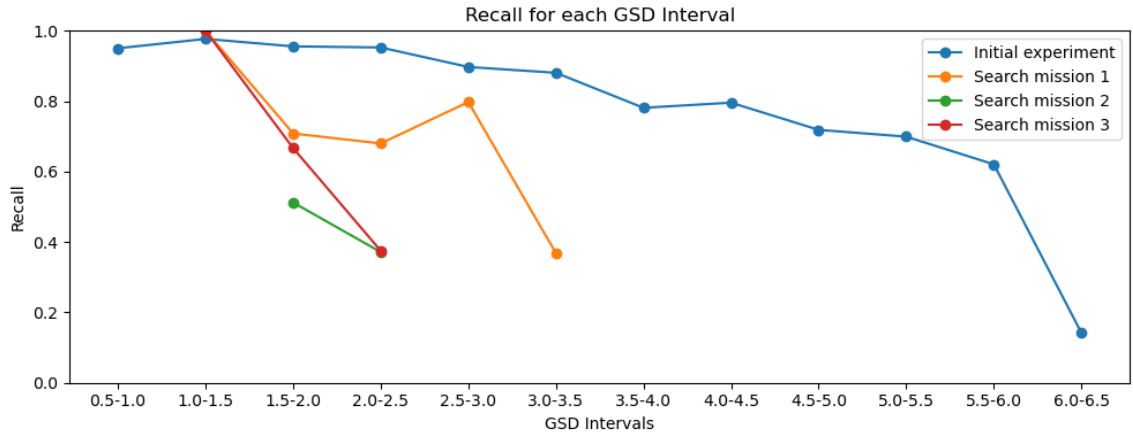

Figure 13. Recall of images created as tiles of images containing people. The initial experiment was operated manually with no height optimization resulting in more GSD intervals than the autonomously operated missions in the search experiment. Nevertheless, it can be seen that the missions' recall follows the general trend of the recall declining with higher GSDs.

*5.3 Validation of the search and detection*

To assess the motion control sensing predicted search accomplishment based on the initial experiment recall, we have compared it to the YOLO recall obtained based on the main experiment images and the first detection of the detected individuals. In image 14 the comparison of the search accomplishment and the YOLO detection rate for the first flight of the first search mission is shown. The predicted search accomplishment and the YOLO detection rate follow the same expected trend of detecting more people through time having a similar increase in detection.

Unfortunately, due to the increasing goal height and the camera resolution, it was not possible to identify each individual in the second, third, and fourth flight to assess the success in the same way. The fifth flight doesn't have enough labels to consider the result reliable so it has been discarded for the same analysis. However, even though the individuals were not identified, the YOLO success can be evaluated based on the number of detected persons shown in Figure 14.

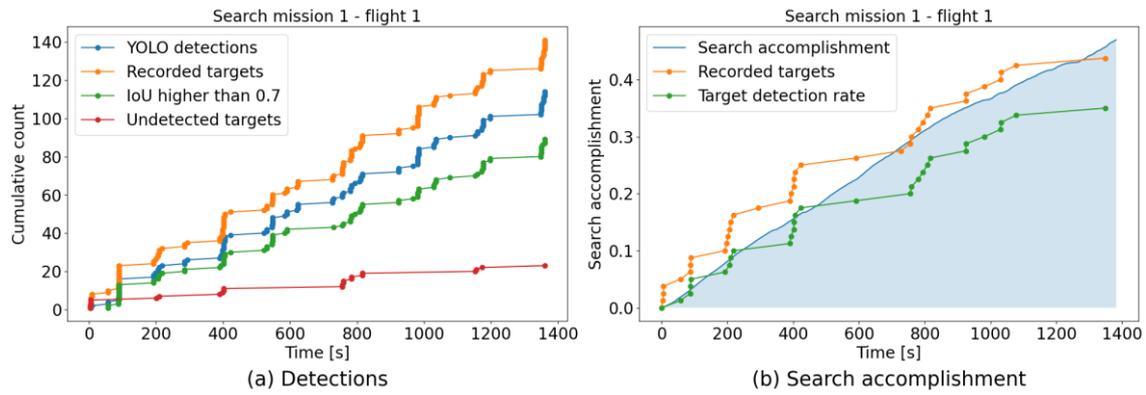

Figure 5.8. The detections and search accomplishment of the first flight in Search mission 1. (a) shows the recorded targets meaning the manually labelled individuals, the YOLO detections of detections with a confidence score higher than 0.5, the detections with an IoU higher than 0.7 and the undetected targets. In (b) the predicted search accomplishment is shown in blue, the recorded manual identifications obtained by images taken in the experiment are shown in orange, while the YOLO detection rate is shown in green. The points show timestamps of images detecting an individual for the first time.

The confusion matrix for search missions 1 and 2 is shown in Figure 15. Search mission 3 consisted of only seven labels, hence it was discarded for further detection analysis. The results suggest that most of the manually detected labels have also been predicted.

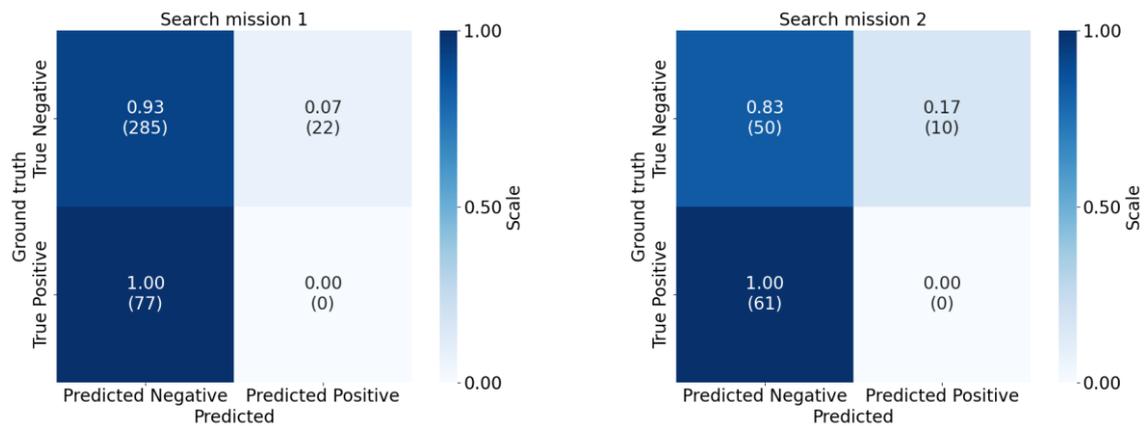

Figure 15. Confusion matrices for all search missions. Most ground truth labels have been detected, while the background area in the ground truth could only be mistakenly predicted as a person causing the maximum percentage being detected as persons. It is important to note that the Search mission 3 only consisted of seven labels, making the results not relevant.

# 6. Discussion and drawbacks

Conducting UAV experiments presents significant challenges due to the unpredictability of environmental conditions. This variability makes it difficult to inform participants well in advance about the confirmed experiment date. Despite these uncertainties, the experiment was successfully carried out with 78 participants in the search and 6 staff members, even though the final date was set only five days before. Weather forecasts have consistently predicted sunny conditions with minimal chance of rain. However, on the experiment day, unexpected fog developed, followed by light rain after the experiment concluded. Additionally, windy conditions can cause additional difficulties, especially in UAV motion control. This has caused unpredictable low quality of the images taken in Search missions 2 and 3 where individuals cannot be identified. In addition to that, the Z30 camera used in the third flight of the Search mission 1 has shown unexpected low quality making it almost impossible to even manually detect individuals.

Managing a considerable large number of participants presents an additional source of logistical challenges. In this study, one of the primary encountered difficulties was data collection, as it relied on participants completing the logs accurately. Despite clearly outlining the required information and providing instructions on how to fill out the forms, analysis of the submitted logs revealed some missing details, such as the names of two participants and the shirt color of multiple individuals. In this case, the missing names did not pose a significant issue, as they could be verified using the participant registration list. However, the shirt color has been shown as a bigger problem since even though multiple individuals have not written any shirt color, there were individuals who have changed the shirt or taken off the jacket during the experiment, but have written only one color. These issues prevent the identification (not detection) of individuals needed to obtain a first detection of each person which is comparable to the search effectiveness $\eta$ calculated in the control framework. Additionally, multiple participants have reported bad signal impacting the real-time map causing issues in tracking the position inside their assigned zone.

# 7. Conclusion

This study presents the simulated SAR mission experiment conducted on Učka mountain, Croatia with the goal of validating the search model, motion control system, as well as gathering additional data for future SAR research. The used motion control consists of the HEDAC algorithm creating a potential field guiding the UAV's direction and MPC for optimizing the flight regime.

The experiment consisted of 150 markers being placed inside three zones ensuring a uniform distribution of participants taking part in the treasure hunt for markers. During the search, three autonomously operated single-UAV search missions were conducted, the first one consisted of three connected flights, while the other two consisted of one flight each.

The results suggest that the probabilistic model of the search has a predicted search accomplishment similar to the manually detected individuals, as well as the YOLO detection rate. By this, the search and motion control systems are validated and show promising results for this method to be used in SAR missions. This study can further be expanded by conducting a search exploration of a simulated SAR mission using multiple UAVs to enhance the efficiency of the search.

**Acknowledgment**

This publication is supported by the Croatian Science Foundation under the project UIP-2020-02-5090.

**Data availability**

The data produced in this research is publicly available on Open Science Framework https://osf.io/kb9e7.